\documentclass[runningheads]{llncs}
\usepackage[T1]{fontenc}
 \usepackage{graphicx}
 \usepackage{multirow}
 \usepackage{booktabs}  
 \usepackage{pifont}
 \usepackage{bbm}
 \usepackage[colorlinks,linkcolor=blue]{hyperref}

\begin{document}

\title{Weakly-supervised High-fidelity Ultrasound Video Synthesis with Feature Decoupling}
\titlerunning{Weakly-supervised High-fidelity US Video Synthesis}
%
\author{Jiamin Liang\inst{1,2,3}\thanks{Jiamin Liang and Xin Yang contribute equally to this work.} \and Xin Yang\inst{1,2,3\star} \and Yuhao Huang\inst{1,2,3} \and Kai Liu\inst{1,2,3} \and Xinrui Zhou\inst{1,2,3} \and Xindi Hu\inst{4} \and Zehui Lin\inst{4} \and Huanjia Luo\inst{5} \and Yuanji Zhang\inst{6} \and Yi Xiong\inst{6} \and Dong Ni\inst{1,2,3}\textsuperscript{(\ding{41})}} 
\institute{
\textsuperscript{$1$}National-Regional Key Technology Engineering Laboratory for Medical Ultrasound, School of Biomedical Engineering, Health Science Center, Shenzhen University, China\\
\email{nidong@szu.edu.cn} \\
\textsuperscript{$2$}Medical Ultrasound Image Computing (MUSIC) Lab, Shenzhen University, China\\
\textsuperscript{$3$}Marshall Laboratory of Biomedical Engineering, Shenzhen University, China\\
\textsuperscript{$4$}Shenzhen RayShape Medical Technology Co., Ltd, China\\
\textsuperscript{$5$}Huizhou Central People's Hospital, Huizhou, Guangdong, China\\
\textsuperscript{$6$}Department of Ultrasound, Luohu People’s Hosptial, Shenzhen, China\\
}
\authorrunning{Liang et al.}


\maketitle
\begin{abstract}
Ultrasound (US) is widely used for its advantages of real-time imaging, radiation-free and portability. In clinical practice, analysis and diagnosis often rely on US sequences rather than a single image to obtain dynamic anatomical information. This is challenging for novices to learn because practicing with adequate videos from patients is clinically unpractical. In this paper, we propose a novel framework to synthesize high-fidelity US videos. Specifically, the synthesis videos are generated by animating source content images based on the motion of given driving videos. Our highlights are three-fold. First, leveraging the advantages of self- and fully-supervised learning, our proposed system is trained in weakly-supervised manner for keypoint detection. These keypoints then provide vital information for handling complex high dynamic motions in US videos. Second, we decouple content and texture learning using the dual decoders to effectively reduce the model learning difficulty. Last, we adopt the adversarial training strategy with GAN losses for further improving the sharpness of the generated videos, narrowing the gap between real and synthesis videos. We validate our method on a large in-house pelvic dataset with high dynamic motion. Extensive evaluation metrics and user study prove the effectiveness of our proposed method. 
\end{abstract}

\section{Introduction}
\label{Introduction}
Ultrasound (US) videos can provide more diagnostic information flow compared to static images, thus being popular in various clinical scenarios.
Sonographers require to learn through abundant US video scans for gaining experiences and improving diagnostic ability.
However, acquisition of plenty of US sequences with teaching and diagnostic significance is unpractical in clinic.
Specifically, scanning high-quality US videos including numerous frames is time-consuming, also, operator- and device-dependent. Besides, some positive cases and rare diseases have a limited amount in clinical practice, which results in their collecting difficulties. 
Hence, synthesizing US videos with high fidelity and dynamic motion is highly desirable to assist in clinical training. 

In the related studies, lots of generative adversarial network (GAN)~\cite{GAN} based approaches have been explored to synthesize medical images~\cite{freehand2017,simulating2018,liang2020synthesis}.
Though synthesizing realistic images, these methods focus on image-level static information, without considering information flow between frames, and thus cannot be directly used in video synthesis task.
To date, several video-based synthesis methods have been proposed, and they can be roughly classified into two types.

\begin{figure}[!t]
\includegraphics[width=\textwidth]{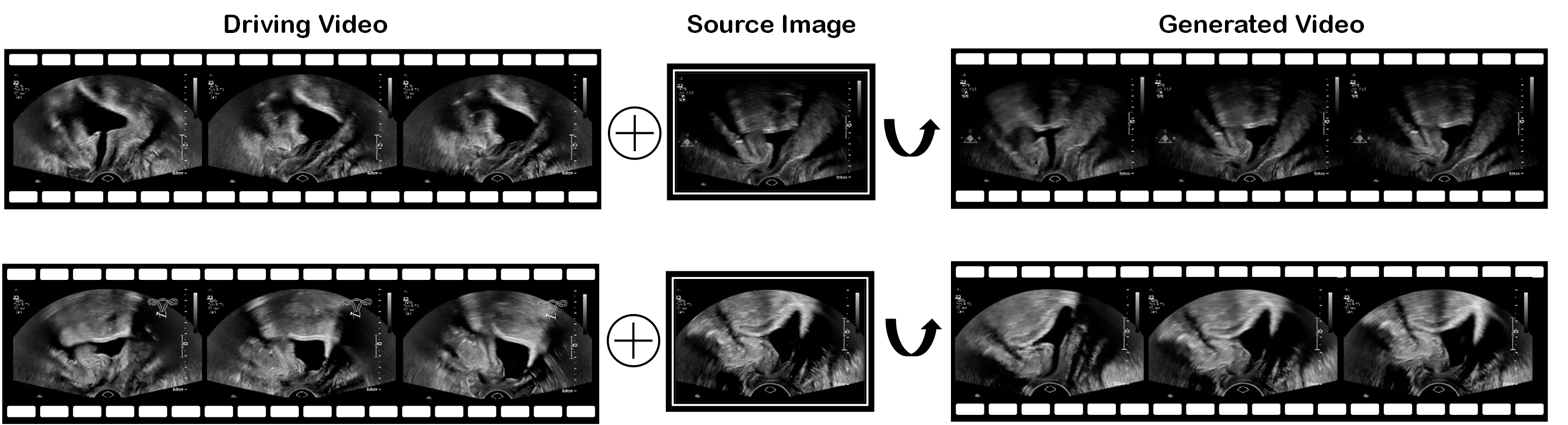}
\caption{Our task is to transfer the motion from the driving video to the static source image, thus obtaining the generated US video.}
\label{fig_task}
\end{figure}

\textbf{Unconditional Video Synthesis (UVS).}
Most UVS methods took random noise as input and learned both content and motion information with high complexity.
The most common UVS approaches set different vectors for learning image content and motion, respectively~\cite{VGAN,TGAN,mocogan}.
However, due to the lack of informative driving signals, a large degree of distortion may occur as generated frames increase or motions get complex.
Thus, these methods cannot handle the long-time or large-movement-range video generations.

\textbf{Conditional Video Synthesis (CVS).}
Compared to UVS, CVS took additional content or motion as input, thus improving the synthesis quality.
Realistic face video simulation based on 3D face model is one focus of CVS research~\cite{zollhofer2018state}. 
Then, inspired by the image-to-image translation framework, a video-to-video system using large-scale data and paired segmentation maps was proposed to synthesize high-resolution and temporally consistent videos~\cite{wang2018video}.
These methods depend on strong priors or annotations, which limits their applicability in medical, especially US, video synthesis tasks.
Most recently, some annotation-free methods were proposed to transfer motions among images by taking static source images and driving videos as the~\textit{condition}.
Specifically, source image and driving video were used as appearance and action information, respectively. 
X2Face~\cite{wiles2018x2face} decomposed identity and pose to synthesize new face by warping static images according to the driving video.
Monkey-Net~\cite{monkeynet} employed unsupervised learning to detect sparse motion-specific keypoints.
Following~\cite{monkeynet}, Siarohin et al.~\cite{fomm} proposed first order motion model (FOMM) to predict keypoint local affine transformations, further improving the generated video quality.

Though the above-mentioned methods have been validated on human pose and face datasets, they are still challenging to synthesize US videos with high fidelity and high dynamic motion.
First, speckle noise in  video may make unsupervised models fail in perceiving anatomical areas, thus causing severe image distortion.
Moreover, US videos usually contain structures of varying size or intensity representation, and patient movements cannot be strictly controlled during US scanning. This will result in very complex and uncertain motion trajectories, making model learning difficult. \par

In this paper, we propose a novel framework for high-fidelity US video synthesis.
The proposed framework animates static source images for video generation by extracting motion information from given driving videos (see Fig.~\ref{fig_task}).
We believe the proposed framework is the first US video synthesis system.
Our contributions are three-fold. First, we leverage weakly-supervised learning to predict the keypoints, thus capturing the complex high dynamic motions in US videos.
Note that we only need few keypoint annotations during training, and the rest can be learned by the model automatically. Second, we carefully design a two-branch architecture to learn content and texture separately, thus simplifying model optimization.
Last, we adopt adversarial learning and GAN losses to further enhance the sharpness of generated frames.
Validation experiments and user study demonstrate the efficacy of the proposed framework.

\section{Methodology}
\label{Method}

\begin{figure*}[t]
	\centering
	\includegraphics[width=1.0\linewidth]{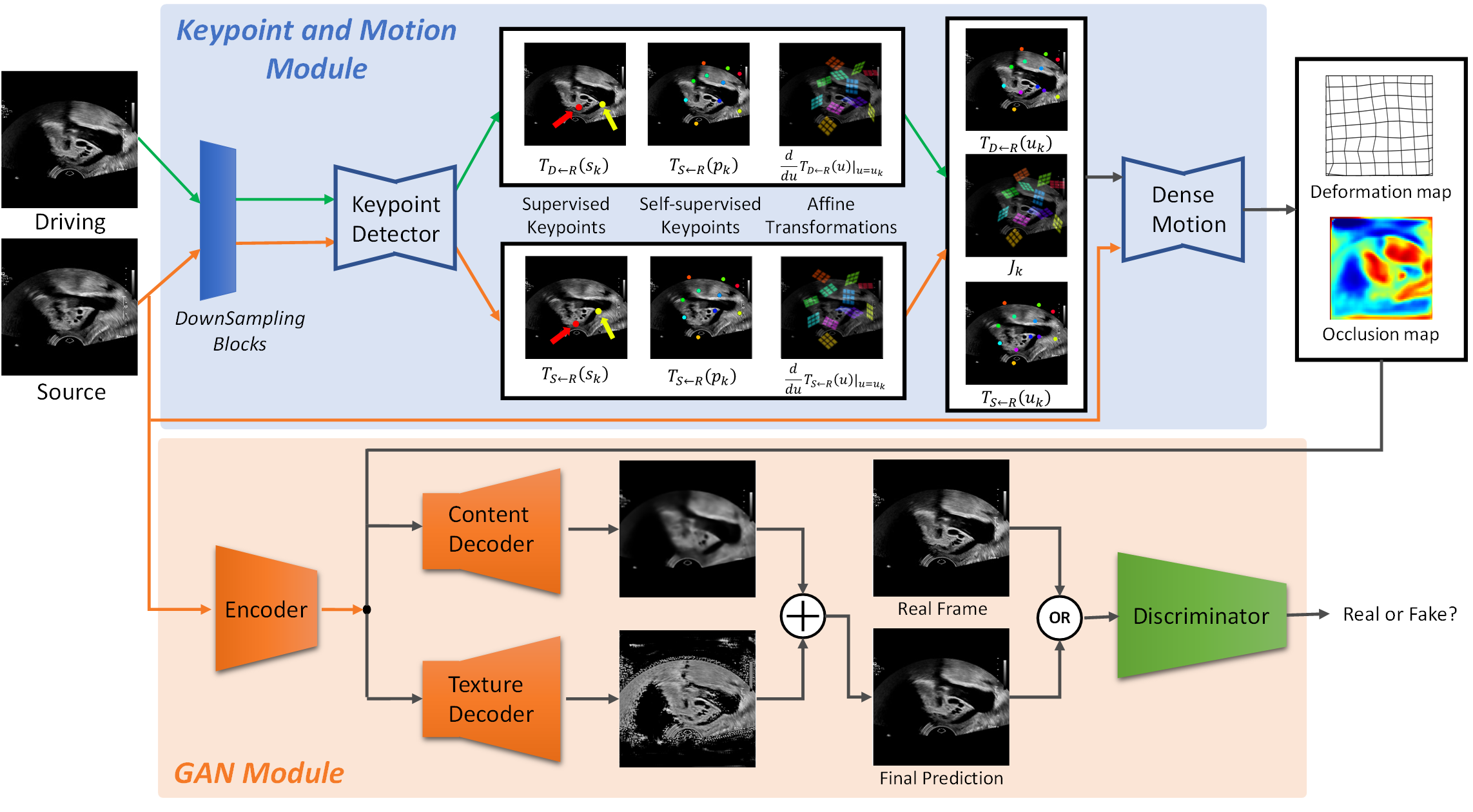}
	\caption{Our proposed framework. In the \textit{supervised keypoints}, red (left) and yellow (right) arrows indicate the lower edge of the symphysis and the bladder neck.}
	\label{fig_framework}
\end{figure*}

Fig.~\ref{fig_framework} shows our proposed framework for high-fidelity US video synthesis. 
Our task is to animate the static source image (\textbf{S}) via the motion information provided by the driving video (\textbf{D}). 
In our proposed system, we first train a keypoint detector to predict the points and their affine transformations in weakly-supervised manner.
Second, a motion prediction network is equipped for estimating the deformation and occlusion maps. Last, a generator with dual-branch decoder and a discriminator are introduced for high-frequency information learning, thus ensuring high-quality video generation.

\subsection{Weakly-supervised Training for Motion Estimation}
The locations of keypoints reflect the relative motion relationship between frames, and thus detecting them can help the model learn the motion transformation.
In this study, based on the pure self-supervised keypoint detection~\cite{fomm}, we further add several supervised keypoints for providing vital anatomical information to benefit the model learning.
Hence, the training of keypoint detection is considered as weakly-supervised learning.
Specifically, as shown in Fig.~\ref{fig_framework}, the keypoint detector uses source and driving frames as input, and locates the keypoints in two ways.
The one way is to learn the keypoints in an self-supervised way through the Thin Plate Splines (TPS) deformations (refer to~\cite{fomm}).
Another way is to detect the anatomical points with manual annotation in a fully-supervised manner. Besides, the affine transformations are also predicted by the detector to provide first-order motion information ($\frac{d}{d u}$ in Fig.~\ref{fig_framework}
) of each detected keypoint.
Similar to FOMM~\cite{fomm}, giving the transformations ($\frac{d}{d u}$ in Fig.~\ref{fig_framework}
) considering the assumed reference frame \textbf{R}, the Jacobian matrix (\textbf{$J_{k}$}) can be obtained to help predict motions between \textbf{S} and \textbf{D}.
$u_{k}$ is the coordinates of $k_{th}$ keypoint.

For self-supervised keypoints learning, the equivariance losses $\mathcal{L}_{\mathrm{eq}}$ in terms of displacements and affine transformations are calculated as Eq.~\ref{eq_loss_eq1} and Eq.~\ref{eq_loss_eq2}.

\begin{scriptsize}
\begin{equation}
\label{eq_loss_eq1}
\mathcal{L}_{\mathrm{eq1}} = \left \| T_{\mathbf{X} \leftarrow \mathbf{R}}\left(p_{k}\right) - T_{\mathbf{X} \leftarrow \mathbf{Y}} \circ T_{\mathbf{Y} \leftarrow \mathbf{R}}\left(p_{k}\right) \right \|_{1},    
\end{equation}
\begin{equation}
\label{eq_loss_eq2}
\mathcal{L}_{\mathrm{eq} 2}=\left\|\mathbbm{1}-\left(\left. {T}'_{\mathbf{X} \leftarrow \mathbf{R}}(p)\right|_{p=p_{k}}\right)^{-1}\left(\left. {T}'_{\mathbf{X} \leftarrow \mathbf{Y}}(p)\right|_{p=T_{\mathbf{Y} \leftarrow \mathbf{R}}\left(p_{k}\right)}\right)\left(\left. {T}'_{\mathbf{Y} \leftarrow \mathbf{R}}(p)\right|_{p=p_{k}}\right)\right\|_{1},
\end{equation}
\begin{equation}
\label{eq_loss_eq}
\mathcal{L}_{\mathrm{eq}} = \mathcal{L}_{\mathrm{eq1}} + \mathcal{L}_{\mathrm{eq2}},
\end{equation}
\end{scriptsize}
where $X$ and $Y$ denote the driving/source frames, and the frames after TPS transformation of $X$. $p_{k}$ denotes the $k_{th}$ coordinates of the self-supervised points. \par

For supervised part, we use the L2 loss to constrain the differences between true and predicted heatmaps, which can be calcultated by:
\begin{equation}
\label{eq_loss_keypoint}
\mathcal{L}_{\mathrm{key}}=\| T_{\mathbf{X} \leftarrow \mathbf{R}}\left(s_{k}\right)- Heatmap \left(s_{k}\right) \|_{2},
\end{equation}
where $s_{k}$ denotes the $k_{th}$ coordinates of the supervised keypoint. 
Then, taking the source image and output of keypoint detector as input, we use the dense motion network in \cite{fomm} to learn the dense deformation fields and occlusion maps. 
Theses maps provide vital indication to tell the model where to focus.

\begin{figure*}[t]
	\centering
	\includegraphics[width=1.0\linewidth]{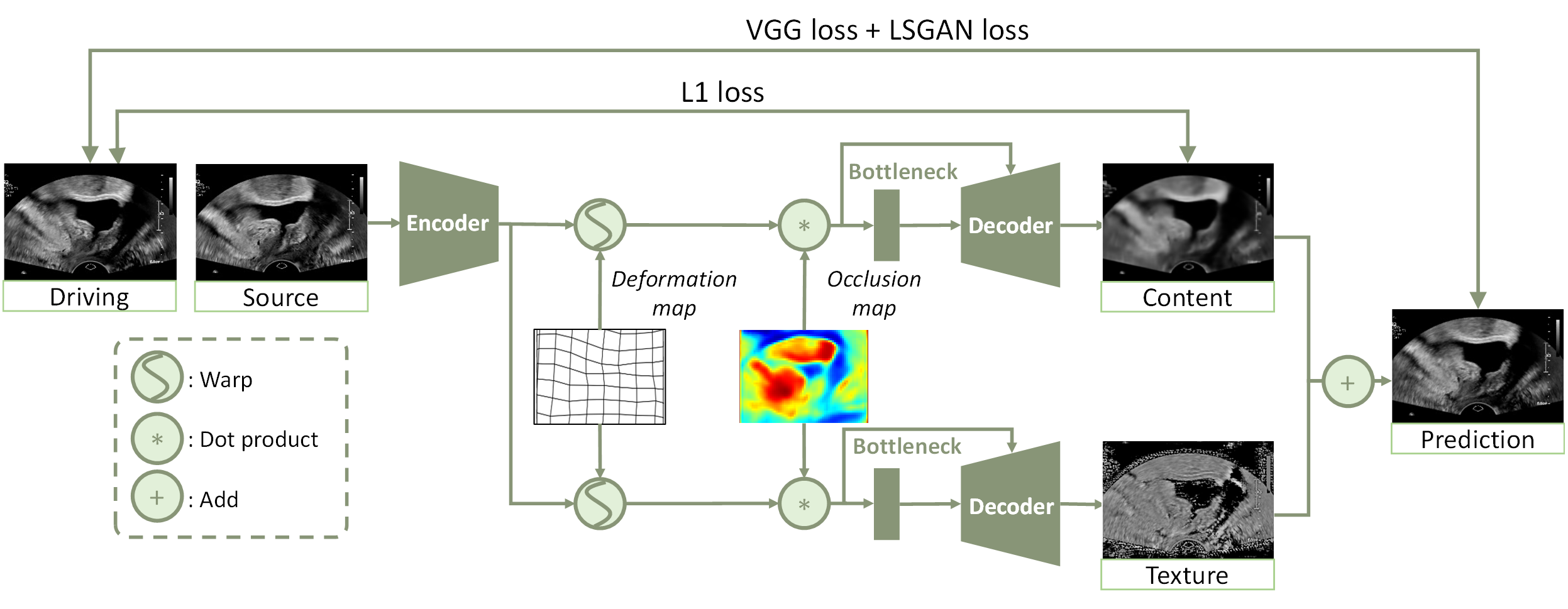}
	\caption{The detailed network structure of generator.}
	\label{fig_generator}
\end{figure*}

\subsection{Dual-Decoder Generator for Content and Texture Decoupling}
\label{sec_dual}
Content and texture are the two vital elements that may influence the visual quality of frame/video.
Most previous studies, taking deformation and occlusion maps as input, adopted a unified path for simultaneously learning content and high-frequency texture.
This architecture design requires the model to decode the highly-coupled features, which may easily cause high-frequency information (i.e., texture) loss, and thus resulting blurry videos. Besides, the lower-size maps may further aggravate this problem in the upsampling stage. In this study, we propose a dual-decoder architecture to decouple the content and texture learning, thus reducing the learning difficulty of the network effectively.

Fig.~\ref{fig_generator} presents the structure of generator. The inputs are the source image, and deformation\&occlusion maps predicted by the dense motion network. 
The intermediate features are then fed to each upsampling layer of the decoder to strengthen the learning of the deformation information, i.e., the two maps. 
The final prediction image is obtained by adding the pixel values of the content and texture images.
The learning mode is driven by the carefully-designed loss functions.
For content learning, we use \textit{L1 reconstruction loss} to restrict pixel-level consistency between the driving image and predicted content.
For texture part, due to unavailability of its ground truth, we adopt feature reconstruction VGG loss~\cite{johnson2016perceptual} to constrain the similarity of driving frame and the final prediction with texture information.
The two losses are calculated on multiple resolutions obtained by \textit{downsample} operations, including 256$\times$256, 128$\times$128, 64$\times$64 and 32$\times$32, which can be written as follows:
\begin{equation}
\label{eq_loss_rec_l1}
L_{rec_{L1}} =\sum_{i=0}^{I}\left\|Down_{i}(\mathbf{D})-Down_{i}(G_{c}(\mathbf{S}))\right\|_{1},	
\end{equation}
\begin{equation}
\label{eq_loss_rec_vgg}
L_{rec_{VGG}} =\sum_{i=0}^{I}\sum_{j=1}^{J}\left\|VGG_{j}(Down_{i}(\mathbf{D}))-VGG_{j}(Down_{i}(G_{f}(\mathbf{S})))\right\|_{1},
\end{equation}
where $Down_{i}$ denotes $i_{th}$ downsample, $VGG_{j}$ is the $j_{th}$ activation layer of VGG network. $G_{c}$ and $G_{f}$ are the content image and final prediction, respectively. 

\subsection{Adversarial Learning and GAN Loss for Sharpness Improvement}
\label{sec_gan_loss}
Though the above designs have provided the informative estimation of motion, content and texture, the synthesis videos still cannot meet the clinical training requirements due to finer details loss when compared to ground truth.
Thus, the adversarial training strategy is further introduced to improve the sharpness of the generated videos.
Specifically, we add a discriminator to judge the differences between reconstructed and real frames, that is, forcing the generator to learn and synthesize more realistic frames to 'cheat' the discriminator. 
The input of the discriminator is generated or real frame and the output is a probability map, predicting the trueness of the frame.
We train the generator and discriminator ($Dis$) using LSGAN loss~\cite{mao2017least} for learning stability (see Eq.~\ref{eq_loss_lsgan_g} and~\ref{eq_loss_lsgan_d}).
\begin{equation}
\label{eq_loss_lsgan_g}
L_{G}^{LSGAN}=E\left[(Dis(G_{f}(\mathbf{S}))-1)^{2}\right]
\end{equation}
\begin{equation}
\label{eq_loss_lsgan_d}
L_{Dis}^{LSGAN}=E\left[(Dis(\mathbf{D})-1)^{2}\right]+E\left[Dis(G_{f}(\mathbf{S}))^{2}\right]
\end{equation}

Further, the feature matching loss~\cite{wang2018high} is adopted to encourage the similar intermediate representation of the discriminator, which can be written as Eq.~\ref{eq_loss_feature}:
\begin{equation}
\label{eq_loss_feature}
L_{feat} =\sum_{i=0}^{I}\left\|Dis_{i}(\mathbf{D})-Dis_{i}(G_{final}(\mathbf{S}))\right\|_{1},
\end{equation}
where $Dis_{i}$ denotes for the $i_{th}$ intermediate outputs of the discriminator.

\section{Experiments and results}
\label{Result}
\begin{table*}[!b]
	\centering
	\caption{Quantitative Result of Compared Methods}\label{tab_quantitative}
	\begin{tabular}{l|l|l|l|l|l|l|l}
		\hline
		
		\multirow{2}{*}{\textbf{Methods}} & \multicolumn{5}{c|}{\textbf{Reconstruction}} & \multicolumn{2}{c}{\textbf{Prediction}} \\
		\cline{2-8}
		
		& L1 Loss$\downarrow$ & FID$\downarrow$ & LPIPS$\downarrow$ & PSNR$\uparrow$ & FVD$\downarrow$ & FID$\downarrow$ & FVD$\downarrow$ \\
		\hline
		M-Net \cite{monkeynet} & 
		0.0416 & 17.78 & 0.0120 & 33.05 &  619.50 &
		20.32 & 737.44 \\
		
		\hline		
		FOMM \cite{fomm} & 
		0.0249 & 15.96 & 0.0065 & 33.51 & 405.63 & 
		18.90 & 658.12 \\
		\hline
		Ours-P & 
		\textbf{0.0222} & 15.66 & 0.0059 & 33.66 & 415.77 & 
		18.66 & 575.90 \\
		\hline
		Ours-PT & 
		0.0224 & 15.39 & 0.0056 & 33.66 & 372.52 & 
		18.41 & 571.26 \\
		\hline
		Ours-PTG & 
		0.0225 & \textbf{14.53} & \textbf{0.0044} & \textbf{33.68} & \textbf{324.95} & 
		\textbf{17.82} & \textbf{552.80}\\
		\hline
	\end{tabular}
\end{table*}

\textbf{Materials and Implementation Details.}
We evaluate on pelvic video dataset using endosonography.
Two keypoints (the lower edge of the symphysis and bladder neck) were manually annotated on each frame by experts using the Pair annotation software package \cite{liang2022sketch} (see Fig.~\ref{fig_framework}). 
Totally 169 videos were collected, with 134 for training and 35 for testing. 
Each video contains 37 to 88 frames, which are resized and padded to 256$\times$256.
We implemented our method in \textit{Pytorch} and trained the system by Adam optimizer for 50 epochs, using a standard PC with four NVIDIA TITAN 2080 GPU. The batch size is 20 and the learning rate is set as 0.0002. The number of self-supervised keypoints is set to 10.
The keypoint detection network and dense motion network employ U-Net \cite{ronneberger2015u} structure with five downsampling and upsampling blocks.
The bottleneck contains the structure of 6 residual blocks with two convolution layers, while the discriminator adopts the structure of PatchGAN~\cite{isola2017image} with four convolution layers.
The weights of losses $\mathcal{L}_{\mathrm{eq}}$, $\mathcal{L}_{\mathrm{key}}$, $L_{rec_{L1}}$, $L_{rec_{VGG}}$, $L_{G}^{LSGAN}$, $L_{Dis}^{LSGAN}$, $L_{feat}$ are 10, 100, 10, 10, 1, 1, 10, respectively.

\begin{figure*}[t]
	\centering
	\includegraphics[width=1.0\linewidth]{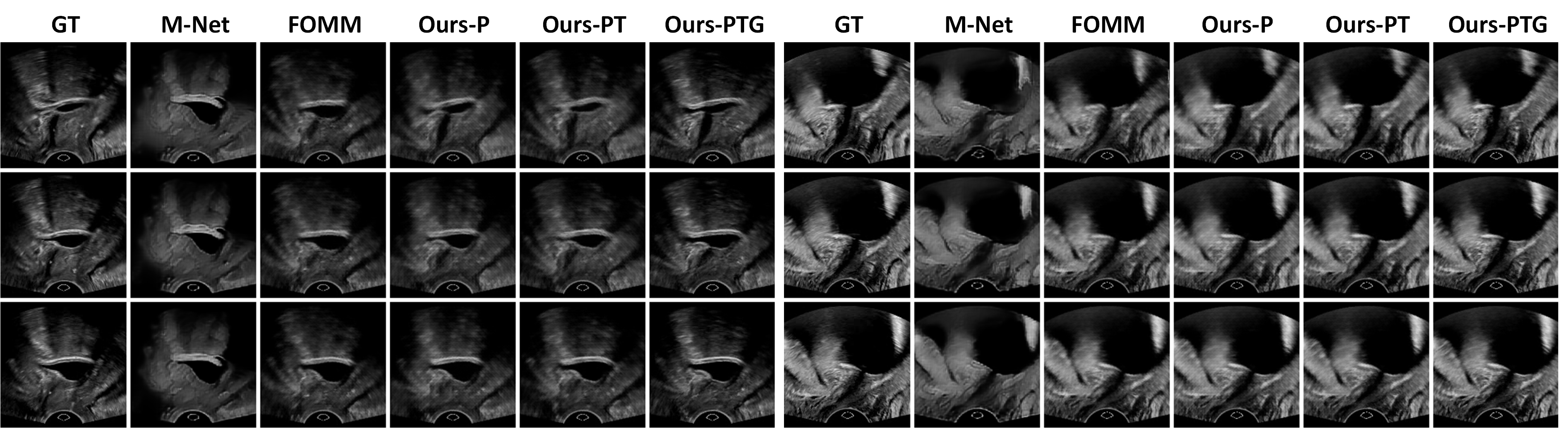}
	\caption{Visualization results of two typical cases for reconstruction task.}
	\label{fig_result_reconstrction}
\end{figure*}

\begin{figure*}[t]
	\centering
	\includegraphics[width=1.0\linewidth]{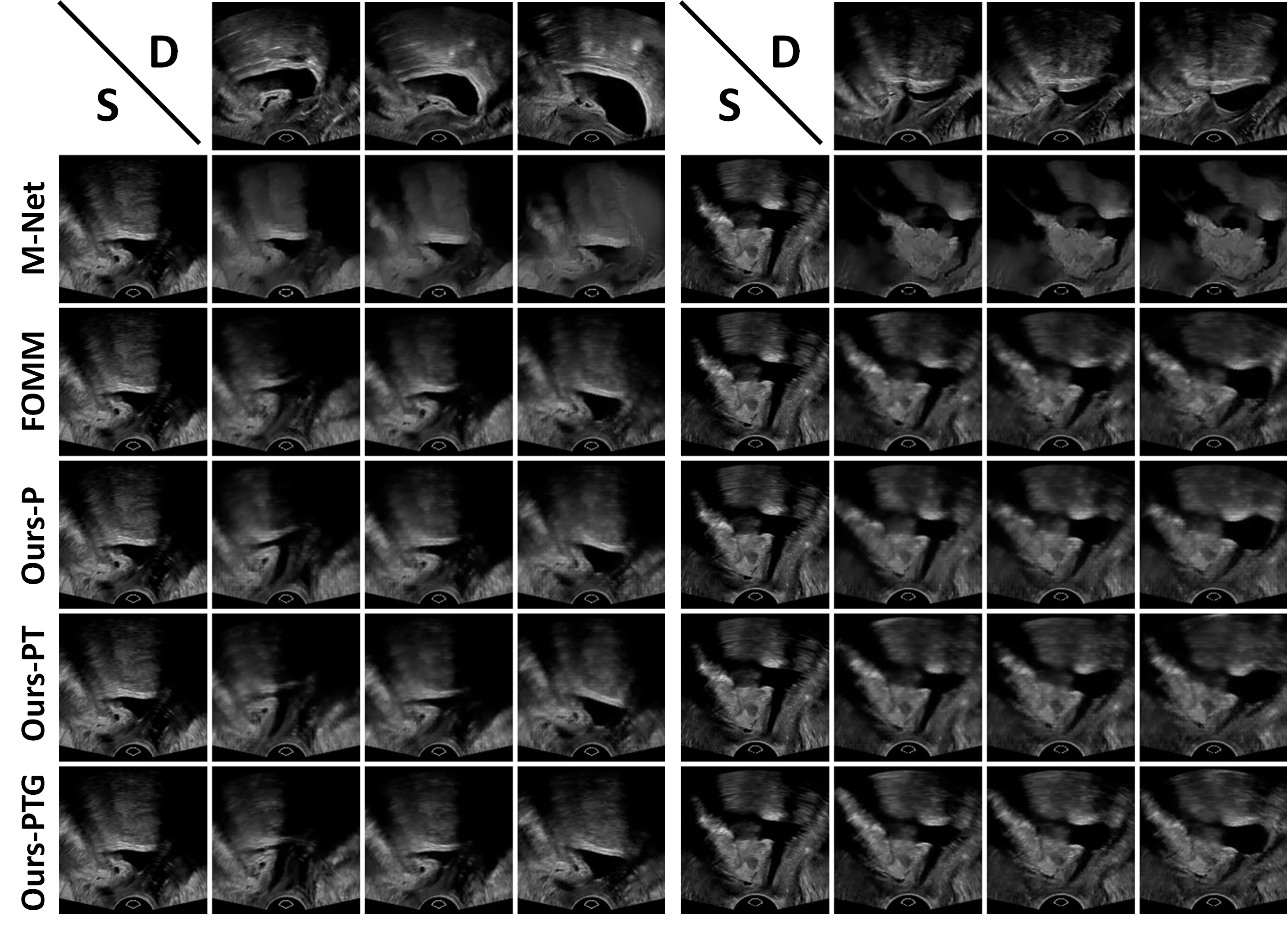}
	\caption{Result visualization for prediction task. D: driving frames; S: source frame.}
	\label{fig_result_prediction}
\end{figure*}

\textbf{Quantitative and Qualitative Analysis.}
We evaluated the performance on two tasks, including reconstruction and prediction. 
For the reconstruction task, we considered the test videos as driving videos and their intermediate frames as source images.
For the prediction task, we randomly choose a video from test set as driving video and an intermediate frame from another test video as source image.
It is noted that the ground truth (GT) in the reconstruction task is the test video itself, while in the prediction task, GT is unavailable.
In this study, five metrics were adopted to evaluate the reconstruction task, including 1) $\mathcal{L}_{1}$ \emph{Loss} for pixel-level absolute distance calculation, 2) \textit{Frechet Inception Distance (FID)}~\cite{FID} for image quality assessment in feature level, 3) \textit{Learned Perceptual Image Patch Similarity (LPIPS)} \cite{LPIPS} for statistics of feature similarity, 4) \textit{Peak Signal to Noise Ratio (PSNR)} for image quality assessment in image level, 5) \textit{Frechet Video Distance (FVD)}~\cite{unterthiner2018towards} for temporal coherence evaluation in video level. 
For the prediction task, only FID and FVD were leveraged, since other metrics required pair GT and synthesis videos.
Fig.~\ref{fig_result_reconstrction} and Fig.~\ref{fig_result_prediction} shows our qualitative results on reconstruction and prediction task, respectively.
Ours-P, ours-PT and ours-PTG denote our ablation studies, including gradually adding keypoint supervision (‘-P’), texture decoder (‘-T’) and GAN loss (‘-G’) to the plain FOMM.
To use the same source image and driving video, ours-PTG achieve the comparable results with GT frames on reconstruction task.
Compared to other FOMM on prediction task, ours-P performs better on the area near the annotated keypoints. Further, with texture enhanced and GAN loss enforced, ours-PTG realizes sharper synthesized frames with high fidelity and consistency between frames.
The quantitative results in Table.~\ref{tab_quantitative} are in accordance with the visualization results.
Our proposed method with keypoint supervision, dual decoder and GAN loss achieves the best performance.

\textbf{User Study.}
To further investigate the quality of synthesis videos, we conducted a user study. Four experienced doctors were asked to rate each giving video in 5 levels. Level 5 means the video looks the most realstic.
Three types of videos (GT, FOMM, ours-PTG) were giving, with each selecting 10 videos. Finally, the average level of the each type was calculated, which was 4.60, 4.15 and 4.65 for videos of GT, FOMM, ours-PTG. The details are presented in Tab.~\ref{tab_user_study}.
With the similar average level of the GT and ours-PTG, we can conclude that the videos synthesized by our method are realistic enough.
\begin{table*}
	\centering
	\caption{The Quantitative Results of User Study}\label{tab_user_study}
	\begin{tabular}{l|l|l|l|l|l}
		\toprule[1pt]
		\textbf{Settings} & \textbf{Doctor 1} & \textbf{Doctor 2} & \textbf{Doctor 3} & \textbf{Doctor 4} & \textbf{Average}\\
		\hline
		\textbf{GT} & 
		4.70 & 4.10 & 4.60 & 5.00 & 4.60 \\
		\hline		
		\textbf{FOMM} & 
		4.60 & 2.40 & 4.60 & 5.00 & 4.15 \\
		\hline
		\textbf{Ours-PTG} & 
		5.00 & 3.90 & 4.80 & 4.90 & 4.65 \\
		\bottomrule[1pt]
	\end{tabular}
\end{table*}

\section{Conclusions}
In this paper, we propose a novel framework for synthesizing high-fidelity US videos to address the challenge of lacking adequate US sequences for training junior doctors.
The videos are synthesized by animating the content in any source images according to the motion of given driving videos.
Extensive experiments on one large pelvic dataset validate the effectiveness of each of our key designs. 
Besides, user study indicates that videos generated by our framework scored close to the real ones, showing the clinical availability of our proposed method.
In the future, we will explore the framework in more challenging datasets to further validate its generality.

\subsubsection{Acknowledgement.} 
This work was supported by the grant from National Natural Science Foundation of China (Nos. 62171290, 62101343), Shenzhen-Hong Kong Joint Research Program (No. SGDX20201103095613036), and Shenzhen Science and Technology Innovations Committee (No. 20200812143441001).

\bibliographystyle{splncs04}
\bibliography{paper1283}

\begin{thebibliography}{10}
\providecommand{\url}[1]{\texttt{#1}}
\providecommand{\urlprefix}{URL }
\providecommand{\doi}[1]{https://doi.org/#1}

\bibitem{GAN}
Goodfellow, I., Pouget-Abadie, J., et~al.: Generative adversarial nets. In:
  NeurIPS. pp. 2672--2680 (2014)

\bibitem{FID}
Heusel, M., Ramsauer, H., et~al.: Gans trained by a two time-scale update rule
  converge to a local nash equilibrium. In: NeurIPS. pp. 6626--6637 (2017)

\bibitem{freehand2017}
Hu, Y., Gibson, E., et~al.: Freehand ultrasound image simulation with
  spatially-conditioned generative adversarial networks. In: Molecular imaging,
  reconstruction and analysis of moving body organs, and stroke imaging and
  treatment, pp. 105--115. Springer (2017)

\bibitem{isola2017image}
Isola, P., Zhu, J.Y., Zhou, T., Efros, A.A.: Image-to-image translation with
  conditional adversarial networks. In: Proceedings of the IEEE conference on
  computer vision and pattern recognition. pp. 1125--1134 (2017)

\bibitem{johnson2016perceptual}
Johnson, J., Alahi, A., Fei-Fei, L.: Perceptual losses for real-time style
  transfer and super-resolution. In: European conference on computer vision.
  pp. 694--711. Springer (2016)

\bibitem{liang2022sketch}
Liang, J., Yang, X., Huang, Y., Li, H., He, S., Hu, X., Chen, Z., Xue, W.,
  Cheng, J., Ni, D.: Sketch guided and progressive growing gan for realistic
  and editable ultrasound image synthesis. Medical Image Analysis  \textbf{79},
   102461 (2022)

\bibitem{liang2020synthesis}
Liang, J., Yang, X., Li, H., Wang, Y., Van, M.T., Dou, H., Chen, C., Fang, J.,
  Liang, X., Mai, Z., et~al.: Synthesis and edition of ultrasound images via
  sketch guided progressive growing gans. In: 2020 IEEE 17th International
  Symposium on Biomedical Imaging (ISBI). pp. 1793--1797. IEEE (2020)

\bibitem{mao2017least}
Mao, X., Li, Q., Xie, H., Lau, R.Y., Wang, Z., Paul~Smolley, S.: Least squares
  generative adversarial networks. In: Proceedings of the IEEE international
  conference on computer vision. pp. 2794--2802 (2017)

\bibitem{ronneberger2015u}
Ronneberger, O., Fischer, P., Brox, T.: U-net: Convolutional networks for
  biomedical image segmentation. In: International Conference on Medical image
  computing and computer-assisted intervention. pp. 234--241. Springer (2015)

\bibitem{TGAN}
Saito, M., Matsumoto, E., Saito, S.: Temporal generative adversarial nets with
  singular value clipping. In: Proceedings of the IEEE international conference
  on computer vision. pp. 2830--2839 (2017)

\bibitem{monkeynet}
Siarohin, A., Lathuili{\`e}re, S., Tulyakov, S., Ricci, E., Sebe, N.: Animating
  arbitrary objects via deep motion transfer. In: Proceedings of the IEEE/CVF
  Conference on Computer Vision and Pattern Recognition. pp. 2377--2386 (2019)

\bibitem{fomm}
Siarohin, A., Lathuili{\`e}re, S., Tulyakov, S., Ricci, E., Sebe, N.: First
  order motion model for image animation. Advances in Neural Information
  Processing Systems  \textbf{32} (2019)

\bibitem{simulating2018}
Tom, F., et~al.: Simulating patho-realistic ultrasound images using deep
  generative networks with adversarial learning. In: ISBI. pp. 1174--1177. IEEE
  (2018)

\bibitem{mocogan}
Tulyakov, S., Liu, M.Y., Yang, X., Kautz, J.: Mocogan: Decomposing motion and
  content for video generation. In: Proceedings of the IEEE conference on
  computer vision and pattern recognition. pp. 1526--1535 (2018)

\bibitem{unterthiner2018towards}
Unterthiner, T., van Steenkiste, S., Kurach, K., Marinier, R., Michalski, M.,
  Gelly, S.: Towards accurate generative models of video: A new metric \&
  challenges. arXiv preprint arXiv:1812.01717  (2018)

\bibitem{VGAN}
Vondrick, C., Pirsiavash, H., Torralba, A.: Generating videos with scene
  dynamics. Advances in neural information processing systems  \textbf{29}
  (2016)

\bibitem{wang2018video}
Wang, T.C., Liu, M.Y., Zhu, J.Y., Liu, G., Tao, A., Kautz, J., Catanzaro, B.:
  Video-to-video synthesis. arXiv preprint arXiv:1808.06601  (2018)

\bibitem{wang2018high}
Wang, T.C., Liu, M.Y., Zhu, J.Y., Tao, A., Kautz, J., Catanzaro, B.:
  High-resolution image synthesis and semantic manipulation with conditional
  gans. In: Proceedings of the IEEE conference on computer vision and pattern
  recognition. pp. 8798--8807 (2018)

\bibitem{wiles2018x2face}
Wiles, O., Koepke, A., Zisserman, A.: X2face: A network for controlling face
  generation using images, audio, and pose codes. In: Proceedings of the
  European conference on computer vision (ECCV). pp. 670--686 (2018)

\bibitem{LPIPS}
Zhang, R., Isola, P., Efros, A.A., Shechtman, E., Wang, O.: The unreasonable
  effectiveness of deep features as a perceptual metric. In: Proceedings of the
  IEEE conference on computer vision and pattern recognition. pp. 586--595
  (2018)

\bibitem{zollhofer2018state}
Zollh{\"o}fer, M., Thies, J., Garrido, P., Bradley, D., Beeler, T., P{\'e}rez,
  P., Stamminger, M., Nie{\ss}ner, M., Theobalt, C.: State of the art on
  monocular 3d face reconstruction, tracking, and applications. In: Computer
  Graphics Forum. vol.~37, pp. 523--550. Wiley Online Library (2018)

\end{thebibliography}
\end{document}